\documentclass{article}

\usepackage[preprint]{neurips_2021}

\usepackage{natbib}
\setcitestyle{numbers,square}
\usepackage{url}
\usepackage{amsmath,amsfonts}
\usepackage{algorithmic}
\usepackage{array}
\usepackage[caption=false,font=normalsize,labelfont=sf,textfont=sf]{subfig}
\usepackage{textcomp}
\usepackage{stfloats}

\usepackage{verbatim}
\usepackage{graphicx}
\def\BibTeX{{\rm B\kern-.05em{\sc i\kern-.025em b}\kern-.08em
		T\kern-.1667em\lower.7ex\hbox{E}\kern-.125emX}}
\usepackage{balance}

\usepackage[hidelinks,
colorlinks=false,]{hyperref}		
\usepackage{booktabs}       
\usepackage{amsfonts}       
\usepackage{amsmath}
\usepackage{nicefrac}       
\usepackage{microtype}      
\usepackage{xcolor}         
\usepackage{graphicx}
\usepackage{wrapfig}
\usepackage{booktabs}
\usepackage{multirow}
\usepackage{diagbox}
\usepackage{caption}
\usepackage{bbding} 
\usepackage{ulem}
\usepackage[ruled,linesnumbered]{algorithm2e}
\usepackage{amsthm}
\definecolor{COLOR}{RGB}{0,0,0}

\title{Differentiable Search of Accurate and Robust Architectures}

\author{%
	Yuwei Ou,
	Xiangning Xie,
	Shangce Gao,
	Yanan Sun,
	Kay Chen TAN,
	Jiancheng Lv \\
	 \\
	 \\
	  \\
	\texttt{ } \\
}

\begin{document}

\maketitle

\begin{abstract}
Deep neural networks (DNNs) are found to be vulnerable to adversarial attacks, and various methods have been proposed for the defense. Among these methods, the adversarial training has been drawing increasing attention because of its simplicity and effectiveness. However, the performance of the adversarial training is greatly limited by the architectures of target DNNs, which often makes the resulting DNNs with poor accuracy and unsatisfactory robustness. To address this problem, we propose DSARA to automatically search for the neural architectures that are accurate and robust after adversarial training. In particular, we design a novel cell-based search space specially for adversarial training, which improves the accuracy and the robustness upper bound of the searched architectures by carefully designing the placement of the cells and the proportional relationship of the filter numbers. Then we propose a two-stage search strategy to search for both accurate and robust neural architectures. At the first stage, the architecture parameters are optimized to minimize the adversarial loss, which makes full use of the effectiveness of the adversarial training in enhancing the robustness. At the second stage, the architecture parameters are optimized to minimize both the natural loss and the adversarial loss utilizing the proposed multi-objective adversarial training method, so that the searched neural architectures are both accurate and robust. We evaluate the proposed algorithm under natural data and various adversarial attacks, which reveals the superiority of the proposed method in terms of both accurate and robust architectures. We also conclude that accurate and robust neural architectures tend to deploy very different structures near the input and the output, which has great practical significance on both hand-crafting and automatically designing of accurate and robust neural architectures.
\end{abstract}

\section{Introduction}\label{Introduction}
\begin{figure}
	\centering
	\includegraphics[width=0.6\textwidth]{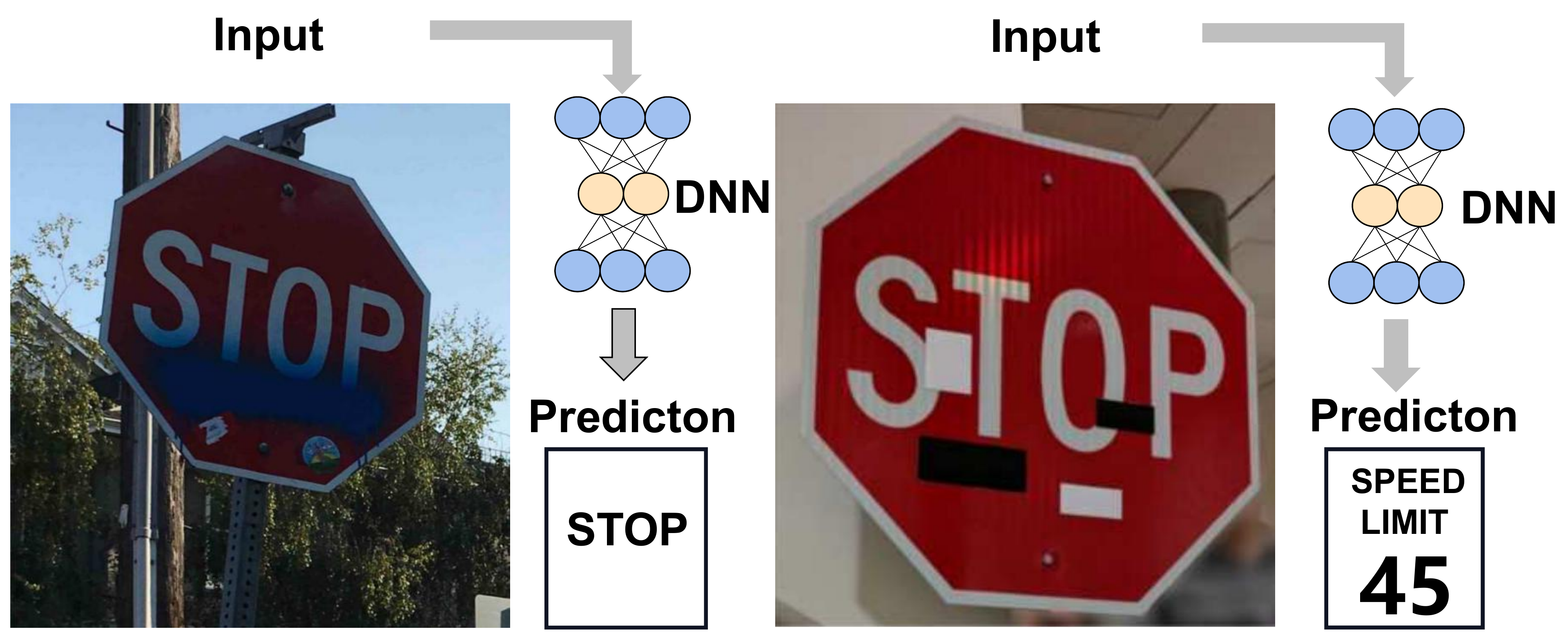}
	\caption{An example of the real-world adversarial attacks. The left image shows real graffiti on a Stop sign, something that most DNNs could recognize correctly. The right image shows a physical perturbation applied to a Stop sign to mimic the real-world scenarios. The image can be identified by humans more clearly than the left image, but is incorrectly recognized by DNNs as the Speed Limit 45 sign.}\label{adversarial}
\end{figure}

Deep neural networks (DNNs) have shown remarkable performance for various real-world applications \cite{krizhevsky2012imagenet,huang2017densely,bahdanau2014neural}. However, DNNs are found to be vulnerable to adversarial attacks \cite{szegedy2013intriguing}, where adding an imperceptible perturbation to the input data can significantly change the prediction result of DNNs. Figure \ref{adversarial} shows a real-world example of adversarial attacks. Concretely, a subtly-modified physical Stop sign can be recognized by the DNNs as a completely different Speed Limit 45 sign, while the image is not significantly changed from a human perspective. The phenomenon implies that DNNs are not robust to small perturbations of their input \cite{szegedy2013intriguing,madry2017towards}, which limits the application of DNNs in security-critical systems such as self-driving cars \cite{eykholt2018robust} and face recognition \cite{sharif2016accessorize}. This is because attackers can maliciously cause safety accidents or create fake identities by crafting imperceptible perturbation to fool the systems. 

After the discovery of this intriguing weakness of DNNs, various methods have been proposed to defend DNNs from adversarial attacks. For example, adversarial training~\cite{madry2017towards,goodfellow2014explaining} enhances the robustness of the DNNs by \textcolor{COLOR}{minimizing the adversarial loss (i.e., loss on adversarial examples that are derived by adding crafted perturbation to the natural data which aims to maximize the prediction error).} Defensive distillation~\cite{papernot2016distillation} is proposed to distill the DNNs so that the final DNNs will not be significantly influenced by the adversarial examples. Data compression~\cite{dziugaite2016study,guo2017countering} is proposed to compress the data before feeding them into the DNNs, so the data after compression will be less likely to be misclassified. Among these methods, the adversarial training is the first proposed method and has been performing the best in enhancing the robustness.

Different from the standard training that trains with natural data directly, the adversarial training trains the network with adversarial examples. Because the adversarial training simulates the fact when suffering from adversarial attacks, the robustness of neural networks can be enhanced through the learning~\cite{goodfellow2014explaining}. \textcolor{COLOR}{However, the robustness resulting from the adversarial training is achieved by adjusting its weight connection during the learning process, which may be limited if the architecture is not well designed in advance. This is because the weight connection is affiliated to the architecture, and only the architecture is determined first, then the weight connections can be adjusted. For example, the experiments of some recent works~\cite{guo2020meets,dong2020adversarially,mok2021advrush} showed that the adversarial training cannot lead to both promising accuracy and robustness if the architectures are poorly designed.} Note that, the accuracy and the robustness of the neural architectures are often measured by the accuracy on natural data and the accuracy on adversarial examples, respectively, which is the convention of the community \cite{goodfellow2014explaining,madry2017towards,tsipras2018robustness}. \textcolor{COLOR}{Therefore, it is necessary to carefully design neural architectures that can obtain high robustness after adversarial training. Please note that the architecture design is not intended to replace the adversarial training. They are two parallel aspects that can be combined to collectively improve the robustness of neural networks.} Unfortunately, designing such neural architectures requires extensive expertise, which is not necessarily available in practice. To address the problem, one feasible solution is to design neural architectures utilizing neural architecture search (NAS) \cite{zoph2016neural}.\par
NAS is a promising technique to design neural architectures automatically. Compared with hand-crafting one, NAS requires less expertise and labor, and can design neural architectures according to the need of specific application scenarios.  As a result, NAS technique has been utilized by some works to search for robust neural architectures which are suitable for the adversarial training. For example, RobNet \cite{guo2020meets} adopts one-shot NAS \cite{bender2018understanding} method, adversarially training a super-network for once and then finding out the robust sub-networks by evaluating each of them under adversarial attacks using the shared weights. ABanditNAS \cite{chen2020anti} introduces an anti-bandit algorithm, searching for robust architectures by gradually abandoning operations that are not likely to make the architectures robust. \textcolor{COLOR}{NADAR~\cite{li2021neural} proposes an architecture dilation method, beginning with the backbone network of a satisfactory accuracy over the natural data, and searching for a dilation architecture to pursue a maximal robustness gain while preserving a minimal accuracy drop. Besides, the most common way is to employ the differentiable search method, and search for robust neural architectures by updating the architectures utilizing some robustness metrics. These robustness metrics include Lipschitz constant~\cite{cisse2017parseval} adopted by RACL \cite{dong2020adversarially}, input loss landscape~\cite{zhao2020bridging} employed by AdvRush~\cite{mok2021advrush}, certified lower bound and Jacobian norm bound proposed and used by DSRNA~\cite{hosseini2021dsrna}. These robustness metrics are all indirect metrics. Specifically, for neural networks, the robustness refers to the prediction ability when the input changes with small perturbations~\cite{szegedy2013intriguing}. Based on this, we divide the robustness metrics into two categories: the direct metrics and the indirect metrics. The direct ones can measure the prediction ability directly, while the indirect ones measure the prediction ability via some proxies. As a result, the robustness metrics aforementioned, including the Lipschitz constant, the input loss landscape, the certified lower bound and the Jacobian norm bound, are indirect metrics. This is because they do not measure the prediction ability directly. To be more specific, the Lipschitz constant is first represented by the spectral norm of the weight matrix, and then the prediction ability is obtained by approximating the spectral norm with the power iteration method~\cite{dong2020adversarially}. The input loss landscape is first quantified by the Frobenius norm of the Hessian matrix, and then the prediction ability is evaluated by approximating the Frobenius norm through the finite difference approximation of the Hessian~\cite{mok2021advrush}. On the contrary, the adversarial loss belongs to the category of the direct metrics, because it measures the loss on the adversarial examples, which directly reflects the prediction ability of neural networks when the input changes with small perturbations. Please note that the natural loss (i.e., loss on natural images) is not a robustness metric because it measures the prediction ability upon the inputs without any changes.}

\begin{figure}
	\centering
	\includegraphics[width=0.6\textwidth]{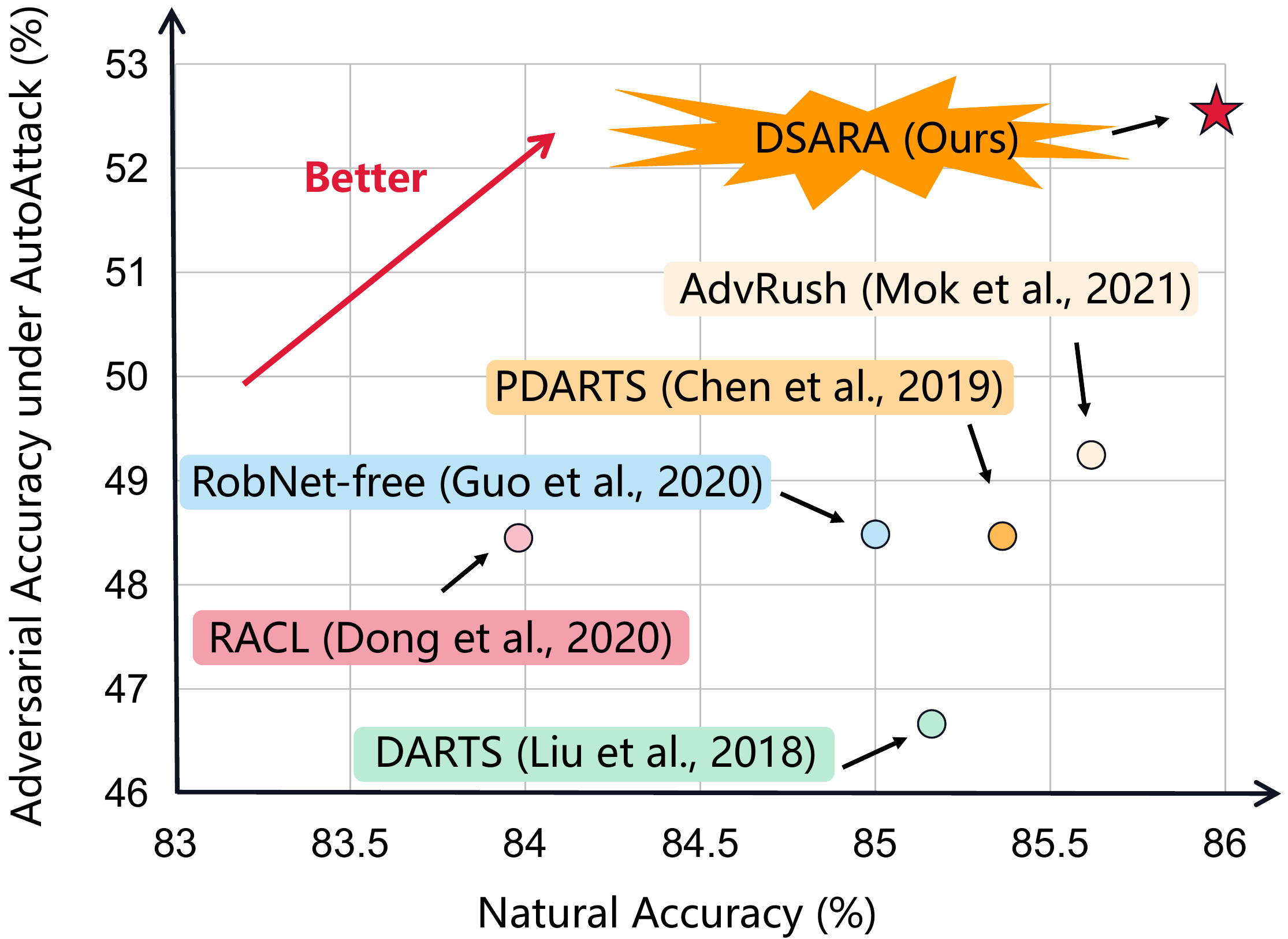}
	\caption{Standard accuracy and adversarial accuracy evaluation results
		on CIFAR-10 for various neural architectures and the neural architecture searched by DSARA. All architectures are adversarially trained using 7-step PGD and the adversarial accuracy is evaluated under AutoAttack. The DSARA architecture achieves the highest standard accuracy and the adversarial accuracy.}\label{compare}
\end{figure}

\textcolor{COLOR}{After a comprehensive investigation of existing methods, we find that} there are still three shortcomings in the existing works. Firstly, most of the existing works directly use the conventional search space \cite{zoph2018learning}. However, it is demonstrated that the adversarial training learns fundamentally different feature representations than the standard training \cite{tsipras2018robustness}. The robust neural architectures which are suitable for the adversarial training may have different overall architecture from the neural architectures which are suitable for the standard training. Consequently, the conventional search space designed for the standard training is no longer suitable for the adversarial training, which results in the low accuracy and the low robustness upper bound of the neural architectures in this search space after adversarial training. Secondly, many of the existing works search for robust architectures through some indirect metrics
related to robustness based on the differentiable search method. However, according to the experimental results of the recent researches~\cite{cisse2017parseval,zhao2020bridging}, the indirect metrics in these works cannot measure the robustness accurately, and usually lead to the weak robustness of the searched neural architectures. The differentiable search utilizing the adversarial training is theoretically a better way to search for robust neural architectures because it directly optimizes the architecture by minimizing the adversarial loss which is in line with our ultimate goal. In spite of the theoretical feasibility of this method, the practice is still largely missing. Thirdly, most of the existing works only take the robustness into consideration during the searching phases, but ignoring the accuracy. However, some works show that there is a trade-off between the accuracy and the robustness~\cite{tsipras2018robustness,zhang2019theoretically}, indicating that taking the robustness as the only optimization objective may result in the low accuracy of the searched neural architectures.

In this paper, we consider the above three problems comprehensively, and propose the Differentiable Search of Accurate and Robust Architectures (DSARA) method. As shown in Figure \ref{compare}, after performing an identical adversarial training procedure, the best neural architecture searched by the proposed DSARA method achieves the highest natural and adversarial accuracies on CIFAR-10. Our contributions can be summarized as follows:

\begin{itemize}
	\item We design a cell-based search space specially for adversarial training. Specifically, we conjecture that the cell structures in different positions of the neural architectures play different roles for both the accuracy and the robustness, so we design Accurate Cell, Robust Cell, and Reduction Cell, then determine the placement of the cells and the proportional relationship of the number of filters through experiments. We conclude that the neural architectures can obtain both the accuracy and the robustness by deploying very different structures in different positions, which has great guiding significance on both hand-crafting and automatically designing of accurate and robust neural architectures.
	\item We propose a two-stage search strategy, searching for accurate and robust neural architectures from the given search space. Specifically, at the first stage, we update the network weights and the architecture parameters alternately to minimize the adversarial loss, which makes full use of the effectiveness of the adversarial training in enhancing the robustness. At the second stage, we update the architecture parameters utilizing the proposed multi-objective adversarial training method. As a result, both the natural loss and the adversarial loss can be explicitly optimized, and the accuracy of the architecture can be further improved.
	\item We conduct extensive experiments to validate the effectiveness of the proposed method under various adversarial attacks on benchmark datasets, which reveals the superiority of the proposed method in terms of both accurate and robust architectures. The experimental results show that the neural architectures searched by our method achieve the highest adversarial accuracy among competitors while the natural accuracy is also outstanding. Meanwhile, the searched architectures are highly transferable to various datasets. We also conduct ablation study to analyze the contribution of each component of the proposed method.
\end{itemize}\par
The remainder of this article is organized as follows. First, the related works are presented in Section \ref{Related Works}. Second, the preliminary knowledge is introduced in Section \ref{Preliminary}. Then, the proposed method are detailed in Section \ref{The Proposed DSARA Method}. Next, the experimental design and the experimental results as well as the analysis are shown in Sections \ref{Experimental Design} and \ref{Experimental Results and Analysis}, respectively. Finally, the conclusions and the future works are given in Section \ref{Conclusion}.
\section{Related Works}\label{Related Works}
\subsection{Adversarial Attacks and Defenses}
Adversarial attack refers to changing the prediction results of DNNs by adding crafted perturbations to the input \cite{szegedy2013intriguing}. Generally, the perturbations are limited to a specific range so that they are imperceptible to human vision systems, while the prediction results can be significantly changed even with a fairly high confidence on the incorrect prediction. Since the discovery of this intriguing weakness of deep neural networks, a lot of adversarial attack methods have been proposed.\par
According to whether the attacker has full access to the target model or not, existing adversarial attack methods can be divided into white-box attacks and black-box attacks. In white-box settings, box-constrained L-BFGS \cite{szegedy2013intriguing} is the first proposed method to perform adversarial attacks. It generates perturbations by optimizing the input to maximize the prediction error. The limitation of the box-constrained L-BFGS is that the computation of the adversarial examples is slow, so it is difficult to use in practice. To address the problem, the fast gradient sign method (FGSM) \cite{goodfellow2014explaining} is proposed to generate adversarial examples by a single step attack that increases the adversarial loss. FGSM accelerates the generation of adversarial examples, but the generated adversarial examples are weaker. To enhance the attack, the basic iterative method (BIM) \cite{kurakin2018adversarial} is proposed to generate adversarial examples in an iterative manner. On this basis, the projected gradient descent (PGD) \cite{madry2017towards} performs random initialization before iteratively updating the perturbation and projects it into a limited range after every iteration. PGD is one of the most popular attack methods since it performs the strongest attack utilizing the local first order information about the network \cite{madry2017towards}. Other white-box attacks include Carlini \& Wagner attacks (C\&W) \cite{carlini2017towards}, DeepFool \cite{moosavi2016deepfool}, the momentum iterative fast gradient sign method (MI-FGSM) \cite{dong2018boosting}, and so on.

In black-box settings, because the attackers have no access to the target model, they generate adversarial examples in completely different ways from the white-box attacks. One commonly used black-box attack method is the transfer-based attacks \cite{papernot2017practical}. The transfer-based attacks generate adversarial examples utilizing a substitute model, which is inspired by the empirical observation that the adversarial examples can transfer between different models. Another commonly used black-box attack method is the score-based attacks \cite{brendel2017decision,chen2017zoo}. The score-based attacks rely on the prediction score of the target model to estimate the gradients, so that the adversarial examples can be generated utilizing the estimated gradients. Recently, AutoAttack \cite{croce2020reliable}, an ensemble of attacks containing both white-box and black-box attacks, becomes popular for robustness evaluation because it is parameter-free, computationally affordable, and user-independent. As a result, in addition to the popular attack methods mentioned before, we also use the AutoAttack in our evaluation stage of the searched neural architectures for reliable comparison.\par
To defend neural networks from adversarial attacks, numerous defense methods have been proposed, among which the adversarial training \cite{madry2017towards,goodfellow2014explaining} is the first proposed method. Although the adversarial training is time consuming, it has been the most popular way so far to help neural networks enhance the robustness \cite{guo2020meets,dong2020adversarially,mok2021advrush,chen2020anti}. To perform the adversarial training, the most widely used method is to replace input data with adversarial examples generated by PGD, for the reason that neural networks adversarially trained using PGD usually generalize to other attacks \cite{madry2017towards}. Defensive distillation \cite{papernot2016distillation} is another method that is effective when defending against some gradient-based attack methods such as FGSM and PGD. It is demonstrated that the defensive distillation disables the gradient calculation of some specific adversarial attack methods \cite{carlini2017towards}, later also known as a method of gradient masking \cite{papernot2017practical}. Obfuscated gradients \cite{athalye2018obfuscated} refer to the case where the constructed defenses necessarily cause gradient masking. The Obfuscated gradients give a false sense of security, because it does not enhance the robustness of the neural architectures themselves, but resists some specific gradient-based attacks \cite{athalye2018obfuscated}. Besides, some other methods such as data compression \cite{dziugaite2016study,guo2017countering}, feature denoising \cite{xie2019feature}, and model ensemble \cite{tramer2017ensemble,pang2019improving} also demonstrate their feasibility. In our work, we notice the fact that the adversarial training, as the most effective defense method, depends on neural architectures. If the neural architecutures are not suitable for the adversarial training, they can only obtain the low accuracy and the low robustness. Our work tackles the problem by automatically designing neural architectures which are good at performing the adversarial training.
\subsection{Neural Architecture Search (NAS)}
NAS is a promising technique which aims to automate the architecture design of DNNs. NAS does not require much expertise and labor when designing neural architecutres according to specific application scenarios, so it becomes more and more popular nowadays \cite{real2019regularized,xu2019pc,hu2022pwsnas}. The NAS pipeline can be categorized into three dimension: search space, search strategy, and performance estimation strategy. \par
The search space defines the candidate neural architectures during the search, which decides the performance upper bound of the searched neural architectures. It is necessary to choose or design the search space according to the specific tasks. There are four types of the commonly used search space: entire-structured \cite{zoph2016neural,brock2017smash} search space, cell-based \cite{zoph2018learning,zhong2018practical} search space, hierarchical \cite{liu2017hierarchical} search space, and morphism-based \cite{wei2016network,chen2015net2net} search space. Firstly, the entire-structured search space builds the neural architectures by stacking a predefined number of nodes, where each node is a layer and performs the specified operation. The entire-structured search space can be hardly used to search for deep neural architectures because the the search cost will increase dramatically with the increase of the depth. Secondly, the cell-based search space constructs the neural architectures by stacking the cells repeatedly, where a cell is composed of some basic layers. The cell-based search space is demonstrated to have transferability, the neural architectures searched from which can achieve competitive results compared to the state-of-the-art hand-crafted models on ImageNet~\cite{krizhevsky2012imagenet}. Thirdly, the hierarchical search space constructs higher-level cells by iteratively incorporating lower-level cells, where the highest-level cell is a single motif representing the full architecture. This method can identify more complex and flexible topologies of the neural architectures. Fourthly, the morphism-based search space designs new neural architectures based on an existing network by inserting identity morphism transformations between the network layers. This method can find deeper or wider models, which are equivalent to the original models. Among these types of search space, the cell-based search space has become the most popular one, and is widely used by many works~\cite{chen2019progressive,xu2019pc,mok2021advrush}. Our work in this paper is also based on the cell-based search space for its effectiveness and transferability. 

The search strategy guides the searching process to search for neural architectures with high performance from the given search space. Popular search strategy includes evolutionary computation-based~\cite{real2019regularized,real2017large} search strategy, reinforcement learning-based~\cite{zoph2016neural} search strategy, and the differentiable search strategy~\cite{liu2018darts,chen2019progressive,xu2019pc}. Early search strategy based on evolutionary computation and reinforcement learning often require thousands of GPU Days, making it expensive to perform NAS in practice. Their main overhead lies in the evaluation for each candidate neural architecture generated during the searching phases. In order to speed up the NAS algorithm \textcolor{COLOR}{by avoiding the architecture evaluation}, the differentiable search strategy is proposed, which relaxes the search space to be continuous, so that the neural architectures could be designed through optimization using gradient descent. The differentiable search strategy does not require any evaluation while searching, so they cost orders of magnitude less computation resources than the aforementioned search strategies. Our work in this paper is based on the differentiable search method for its searching efficiency, and we will introduce it in detail in Section~\ref{Preliminary}.

The performance estimation refers to the process of evaluating models generated during the search when needed. One early performance estimation strategy is to train and evaluate every neural architecture \cite{zoph2016neural,real2017large}, which is fairly time-consuming because there are thousand of neural architectures that need to be evaluated. To accelerate the performance estimation process, several techniques have been proposed. \textcolor{COLOR}{The performance predictor is one of the performance estimation techniques, which even leads to a new research direction called predictor-based NAS~\cite{dudziak2020brp,chen2021contrastive,tang2020semi} for its popularity. In the predictor-based NAS, the performance predictor is used to map the neural architectures to their performance. Thus, the predictor-based NAS is very efficient because it can directly estimate the performance of architectures without training. The limitation is that the predictor-based NAS requires sufficient architecture-performance pairs to train the performance predictor, otherwise the performance predictor is unable to make accurate predictions. For example, E2EPP~\cite{sun2019surrogate} required 1,000 training data to help with the prediction, which is time-consuming. As a result, despite the efficiency of the performance predictor, the predictor-based NAS is still expensive due to the training data collection process.} \textcolor{COLOR}{Another technique is to evaluate neural architectures using proxy tasks~\cite{zhou2020econas} instead of full training. Recently, some low/zero-cost proxies~\cite{abdelfattah2021zero,xu2021knas} are proposed, which also leads to a new research direction called low/zero-cost NAS. In the low/zero-cost NAS, the low/zero-cost proxy tasks are utilized to evaluate the neural architectures without training using random initialized models. For examples, KNAS~\cite{xu2021knas} adopted the gram matrix of gradients as the zero-cost proxy to evaluate the quality of architectures. Abdelfattah et al.~\cite{abdelfattah2021zero} studied on several zero-cost proxies motivated by the pruning literature, such as grad norm (computed by Euclidean norm of the gradients) and synflow (which is simply the product of all parameters in the network). Although the zero-cost NAS can speed up the performance evaluation process by several orders of magnitude compared with the `train-then-test' paradigms, the method cannot generalize well to different datasets and often makes inaccurate evaluation~\cite{ning2021evaluating}. Besides, the weight sharing \cite{pham2018efficient} is also a technique} to accelerate the performance estimation process. In the weight sharing technique, all child models share parameters so that it is no need to train each model separately. Meanwhile, the weight sharing technique \textcolor{COLOR}{requires neither extra training data needed by the performance predictor nor the proxy tasks designed to replace the full training.} Owing to these advantages, the weight sharing technique has been the most popular performance estimation strategy~\cite{ning2020discovering,hu2022pwsnas}, and its efficiency and reliability are proved by practice. The differentiable search strategy mentioned before and adopted by our work essentially utilizes the weight sharing technique as well.

\section{Preliminary}\label{Preliminary}
The proposed algorithm is designed based on the differentiable search method for its searching efficiency. For the better understanding, it is \textcolor{COLOR}{introduced} in detail in this section.\par
In the differentiable search method, a cell can be represented by a directed acyclic graph (DAG) consisting of $N$ ordered nodes. Each node $x^{i}$ is a latent representation (e.g., a feature map in convolutional networks), and each directed edge $(i, j)$ is associated with some operation $o^{(i,j)}$ that applies to $x^{i}$. Considering the case of convolutional cells, each cell includes two input nodes coming from the outputs of previous two cells, while the output of the cell is obtained by concatenating all the intermediate nodes. Each intermediate node is computed via Eq. (\ref{eq1}).

\begin{equation}\label{eq1}
x^i = \sum_{j<i}o^{(i,j)}(x^{(j)})
\end{equation}

In order to optimize the architecture utilizing the gradient descent, the method of continuous relaxation is introduced to relax the categorical choice of a particular operation by applying softmax to all possible operations, which can be formulated as Eq. (\ref{eq2}),

\begin{equation}\label{eq2}
\bar{o}^{(i,j)}=\sum_{o\in \mathcal{O}}\frac{exp(\alpha_{o}^{(i,j)})}{\sum_{o'\in \mathcal{O}}exp(\alpha_{o'}^{(i,j)})}o^{(i,j)}(x^{(i)})
\end{equation}
where $o^{(i,j)}$ is called the mixed operation, $o^{(i,j)}(x^{(i)})$ denotes an optional operation applied to $x^{(i)}$, which comes from a set of candidate operations $O$ (e.g., convolution, pooling, and identity). The operation mixing weights for a pair of nodes $(i, j)$ are parameterized by a vector $\alpha^{(i,j)}$, which is called the architecture parameters. 

After continuous relaxation, both the architecture parameters $\alpha$ and the network weights $\omega$ can be optimized using gradient descent. Thus, the differentiable search method searches for the ideal neural architecture by solving the bi-level optimization problem represented by Eq. (\ref{eq3}),

\begin{equation}\label{eq3}
\begin{cases}
\begin{split}
\underset{\alpha}{min}\quad &\mathcal{L}_{val}(\omega^{*}(\alpha), \alpha) \\
s.t. \quad &\omega^{*}(\alpha)=argmin_{\omega}\ \mathcal{L}_{train}(\omega,\alpha)
\end{split}
\end{cases}
\end{equation}
where the upper-level optimization aims at finding architecture parameters $\alpha$ to minimize the validation loss $\mathcal{L}_{val}(.)$, while the network weights $\omega^{*}$ are derived by minimizing the training loss $\mathcal{L}_{train}(.)$ in the lower-level optimization.\par
At the end of search, a discrete architecture is obtained by replacing each mixed operation with the operation that has the largest architecture parameter by Eq. (\ref{eq4}).
\begin{equation}
o^{(i,j)}=argmax_{o\in \mathcal{O}}\ \alpha_{o}^{(i,j)}\label{eq4}
\end{equation}

\section{The Proposed DSARA Method}\label{The Proposed DSARA Method}
In this section, the proposed DSARA method is documented. First the framework of the proposed DSARA method is presented in Subsection \ref{DSARA Overview}. Then the designed search space is detailed in Subsection \ref{Search Space}. Next the search strategy and the multi-objective adversarial training method proposed by our work are introduced in Subsections \ref{Two-Stage Search Strategy} and \ref{Multi-objective Adversarial Training}, respectively. For in-depth understanding, we also make an analysis to the proposed method in Subsection \ref{Method Analysis}.
\subsection{DSARA Overview}\label{DSARA Overview}	
\begin{algorithm}
	\SetAlgoLined
	\caption{The Framework of DSARA Method}\label{Algorithm_overview}
	\KwIn{$E$ = total number of epochs for search\\
		\qquad \ \ \ $E_{so}$ = number of epochs for the first-stage\\ \qquad \ \ \ optimization\\}
	\KwOut{$f^{*}_{A}$ = final architecture}
	Determine the \textbf{search space with the proposed design}\\
	Initialize the supernet according to the search space based on the differentiable search method\\
	\For(\tcp*[f]{\textbf{\textrm{Two-stage search strategy}}}){$i\leftarrow 1$ \KwTo $E$}{
		\uIf{$i\le E_{so}$}{Perform the first-stage optimization for the supernet using the normal adversarial training method}
		\Else{Perform the second-stage optimization for the supetnet using \textbf{the proposed multi-objective adversarial training method}}
	}
	Derive $f^{*}_{A}$ through discretization rule of the differentiable search strategy from the supernet
\end{algorithm}
The framework of the proposed DSARA method is presented in Algorithm \ref{Algorithm_overview}. Specifically, to begin with, the total number of epochs for search and the number of epochs for the first-stage optimization are given, which divides the searching process into two different phases. Meanwhile, the search space is determined with the proposed design (line 1), and the supernet is constructed according to the designed search space (line 2) for the differentiable search. Next, the searching process is performed by the proposed two-stage search strategy (lines 3 to 9). Specifically, if the current number of epochs is less than the given number of epochs for the first-stage optimization, the first-stage optimization is performed to update the supernet using the normal adversarial training method (line 5). Otherwise, the second-stage optimization is performed to update the supernet using the proposed multi-objective adversarial training method (line 7). Once the search is over, the final neural architecture is discretized from the supernet (line 10).\par
The innovation points of the proposed DSARA method lie in the sophisticated designed search space, the proposed two-stage search strategy, and the proposed multi-objective adversarial training method (they are also highlighted in bold in Algorithm \ref{Algorithm_overview}). They will be introduced in detail in the following subsections.\par

\subsection{Search Space}\label{Search Space}
As mentioned before, the conventional search space \cite{zoph2018learning} designed for the standard training, aiming at achieving the best accuracy, cannot perform well for the adversarial training. This is because the neural architectures which are suitable for the adversarial training require different overall architecture. In order to design a search space for the adversarial training, we should first consider the characteristics of accurate and robust neural architectures. However, both the research of NAS and robust neural networks are emerging topics, there is rare knowledge can be directly explored.

We are aware of a very recent work \cite{huang2021exploring} which explores the architectural ingredients of adversarially robust deep neural networks. It experimentally concludes that the depth and the width of the neural architectures in different positions have dissimilar effects on the accuracy and the robustness. Motivated by this, we have also performed a huge number of experiments, based on which we further make the conclusion shown in Proposition~\ref{as}, which would also be experimentally verified in Subsection~\ref{Architectural Ingredients}. In the following, we will first justify the limitation of the conventional cell-based search space popularly used by existing related works, in response to Proposition~\ref{as}. After that, we will introduce the designed search space in detail.

\newtheorem{Proposition}{Proposition}
\begin{Proposition}\label{as}
	The cells in different positions of the overall architecture may have different effects on the accuracy and the robustness, and the accuracy and the robustness of the neural architectures can be improved simultaneously by placing different cells in different position.
\end{Proposition}

In particular, the commonly used cell-based search space refers to the one first proposed in NASNet~\cite{zoph2018learning}, and then popularized by the famous DARTS algorithm. Currently, almost all differentiable NAS methods adopt this encoding strategy~\cite{chen2019progressive,xu2019pc,dong2020adversarially,mok2021advrush}. Specifically, the cell-based search space only designs two kinds of computation cells named Normal Cells and Reduction Cells, which play the role of enhancing accuracy and reducing data dimension for improving efficiency, respectively. Based on the design, the overall architecture is constructed by stacking multiple Normal Cells to increase the accuracy as much as possible, and rare Reduction Cells to avoid the invalid data dimension. Clearly, the resulting architecture \textcolor{COLOR}{is mainly composed of the same Normal Cells, and the architectures matching Proposition~\ref{as} (such as the architectures that employ the separable convolutions near the input but employ the dilated convolutions near the output) are not contained in the search space, which limits the improvement of accuracy and robustness.} 

Given the above analysis, \textcolor{COLOR}{we retain the Reduction Cell while replacing the single type of Normal Cell with Accurate Cell and Robust Cell, aiming to enlarge the search space to include the neural architectures matching Proposition~\ref{as}.} Consequently, there are three types of cells designed in the proposed search space: Accurate Cell, Robust Cell, and Reduction Cell. In particular, the Accurate Cells and the Robust Cells both return a feature map of the same dimension but can be placed in different positions to take the corresponding effects of accurate and robustness, as concluded in Proposition~\ref{as}. While the Reduction Cells return a feature map where the feature map height and width is reduced by a factor of two, playing the same role as common. With this design, the learned cells could also be stacked to form a full network, resulting the scalability of the designed search space. Based on this, we can construct deep neural architectures while the number of the candidate neural architectures in the search space will not increase. With the designed computation cells in the proposed search space, in the following, we will introduce the details about how to determine their particular position, i.e., the placement order, to achieve both the accuracy and the robustness.

Particularly, the Reduction Cells are placed at one-third and two-thirds of the overall architecture. In the rest of the architecture, the Accurate Cells are placed before the second Reduction Cell, while the Robust Cells are placed after the second Reduction Cell. Meanwhile, instead of the common heuristic \cite{zoph2018learning} that doubling the number of filters in the output whenever the spatial activation size is reduced, we double the number of filters at the first Reduction Cell, and keep the number of filters unchanged at the second Reduction Cell. The effectiveness of this design will be verified in Subsection~\ref{Ablation Study}, \textcolor{COLOR}{and the naming rules (i.e., how to determine which cells are for accuracy and which cells are for robustness) are explained in Subsection~\ref{Architectural Ingredients}.}

\begin{figure}
	\centerline{\includegraphics[width=0.45\textwidth]{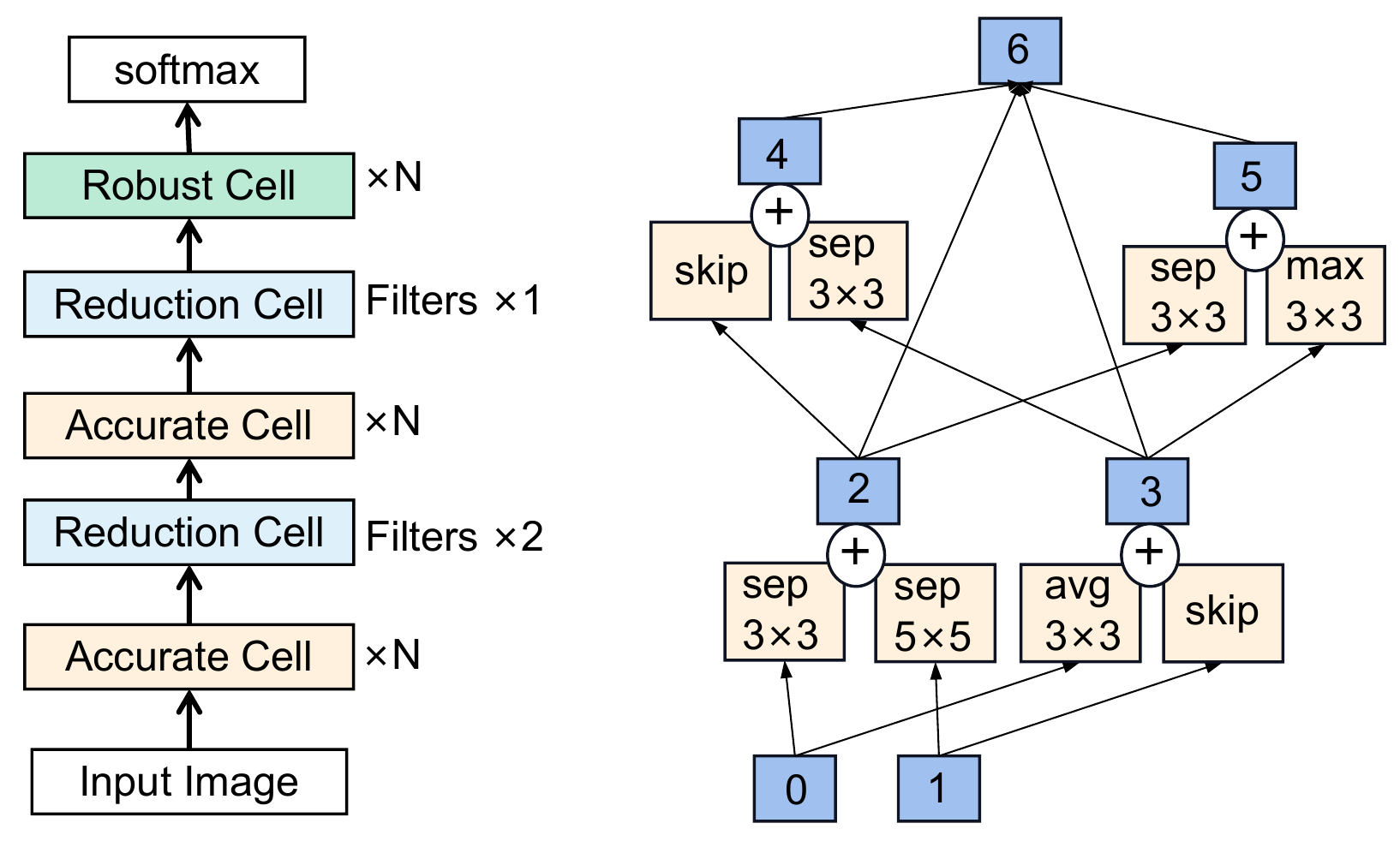}}
	\caption{An example of the proposed search space for CIFAR-10. LEFT: the full outer structure. RIGHT: cell example.}
	\label{fig_search_space}
\end{figure}

For the better understanding of the proposed search space, an example for CIFAR-10 is shown in Figure \ref{fig_search_space}. On the one hand, the left figure shows the full outer structure of the neural architecture. Specifically, the neural architectures are stacked by Accurate Cells, Robust Cells, and Reduction Cells in a determined placement order and a determined proportional relationship of the number of filters as mentioned before. The input data passes through the cells successively, and the softmax function is applied to the output of the final cell to generate prediction results. On the other hand, the right figure shows a particular cell example. Specifically, each cell is composed of 7 nodes and multiple optional operations, including 3×3 and 5×5 separable convolutions, 3×3 and 5×5 dilated separable convolutions, 3×3 max pooling, 3×3 average pooling, and identity.

By designing such a search space, the neural architectures relax the limitation that the cells near the input and the output must be the same. The experimental results show that the accuracy and the robustness upper bounds of the neural architectures from the search space get improved simultaneously, which in turn proves the correctness of the Proposition \ref{as}.

\subsection{Two-Stage Search Strategy}\label{Two-Stage Search Strategy}
\begin{figure*}
	\includegraphics[width=\textwidth]{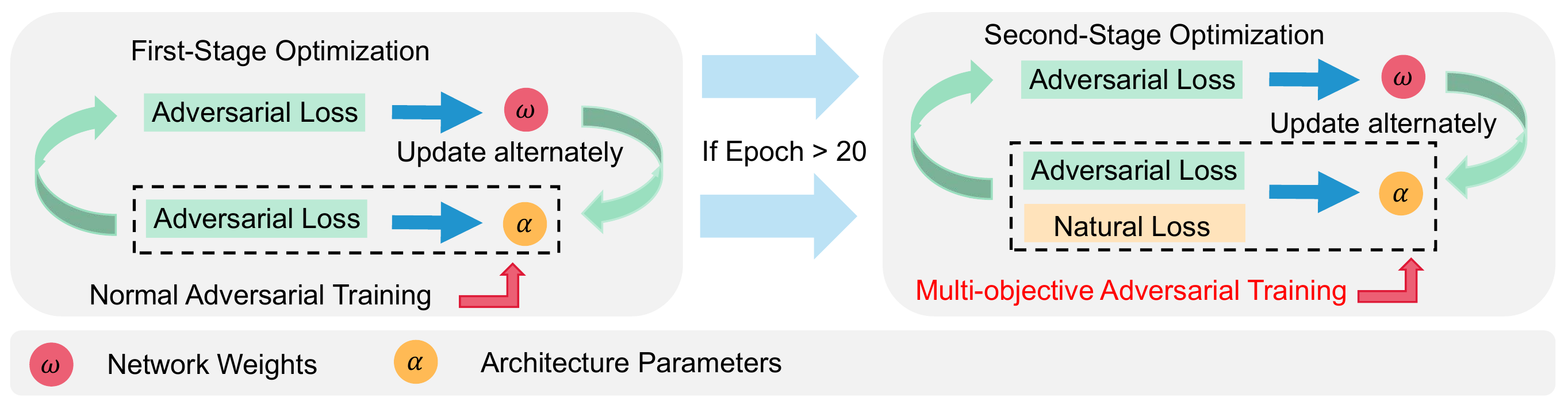}
	\caption{An illustration of the proposed two-stage search strategy.}\label{two-stage}
\end{figure*}
In order to search for both accurate and robust architectures efficiently, we propose a gradient-based two-stage search strategy for the architecture search, performing different forms of the adversarial training at different stages, and the two stages have different effects on enhancing the accracy and the robustness. The pseudo code of the proposed two-stage search strategy is shown in Algorithm \ref{Algorithm1}.

\begin{algorithm}
	\SetAlgoLined
	\caption{Two-Stage Search Strategy}\label{Algorithm1}
	\KwIn{$E$ = total number of epochs for search\\
		\qquad \ \ \ $E_{so}$ = number of epochs for the first-stage optimization\\
		\qquad \ \ \ $\omega_{k}$ = network weights after the iteration $k$\\
		\qquad \ \ \ $\alpha_{k}$ = architecture parameters after the iteration~$k$\\
		\qquad \ \ \ $f_{super}(\omega_{0}, \alpha_{0})$ = supernet initialized from the\\
		\qquad \ \ \ search space}
	\KwOut{$f_{super}(\omega_{E}, \alpha_{E})$ = supernet after optimization }
	\For{$i\leftarrow 1$ \KwTo $E$}{
		\uIf(\tcp*[f]{\textrm{First Stage}}){$i\le E_{so}$}{Update $\omega_{i}$ using $\nabla_{\omega}\mathcal{L}_{train}^{adv}(\omega_{i-1}, \alpha_{i-1})$ \\Update $\alpha_{i}$ using $\nabla_{\alpha}\mathcal{L}_{val}^{adv}(\omega_{i}, \alpha_{i-1})$ }
		\Else(\tcp*[f]{\textrm{Second Stage}}){Update $\omega_{i}$ using $\nabla_{\omega}\mathcal{L}_{train}^{adv}(\omega_{i-1}, \alpha_{i-1})$ \\
			Update $\alpha_{i}$ to minimize $\mathcal{L}_{val}^{std}(\omega_{i}, \alpha_{i-1})$ and $\mathcal{L}_{val}^{adv}(\omega_{i}, \alpha_{i-1})$ simultaneously using the proposed multi-objective adversarial training method
			
		}
	}
\end{algorithm}
At the first stage, we concentrate on improving the robustness of the architecture because we take the robustness as the most important property of the neural architectures. Gennerally, the accuracy of the robust neural architectures will not be too worse. So we alternately update the networks weights and the architecture parameters utilizing single-objective adversarial training to minimize the adversarial loss so that the architecture enhances the robustness against the adversarial examples. The single-objective search process can be formulated as Eq. (\ref{eq5}),
\begin{equation}\label{eq5}
\begin{cases}
\begin{split}
\underset{\alpha}{min}\quad &\mathcal{L}_{val}^{adv}(\omega^{*}(\alpha), \alpha) \\
s.t. \quad &\omega^{*}(\alpha)=argmin_{\omega}\ \mathcal{L}_{train}^{adv}(\omega,\alpha)
\end{split}
\end{cases}
\end{equation}
where $\omega$ denotes the network weights, $\alpha$ denotes the architecture parameters, $\mathcal{L}^{adv}_{train}(.)$ and $\mathcal{L}^{adv}_{val}(.)$ denote the adversarial loss on the training set and the validation set, respectively.\par
At the second stage, we take both the accuracy and the robustness into consideration. Our goal is to further improve the accuracy and the robustness of the neural architectures while the robustness is at least stronger than after the first-stage optimization. So we update the architecture parameters utilizing the multi-objective adversarial training method proposed by our work. The method  minimizes both the natural loss and the adversarial loss in the upper-level optimization to improve the accuracy and the robustness of the architectures simultaneously, the details of which will be introduced in Subsection \ref{Multi-objective Adversarial Training}. We keep the lower-level optimization unchanged because our goal is to search for accurate and robust neural architectures after the normal adversarial training. The multi-objective search process can be formulated as Eq. (\ref{eq6}),
\begin{equation}\label{eq6}
\begin{cases}
\begin{split}
\underset{\alpha}{min}\quad &(\mathcal{L}_{val}^{std}(\omega^{*}(\alpha), \alpha),\ \mathcal{L}_{val}^{adv}(\omega^{*}(\alpha), \alpha)) \\
s.t. \quad &\omega^{*}(\alpha)=argmin_{\omega}\ \mathcal{L}_{train}^{adv}(\omega,\alpha)
\end{split}
\end{cases}
\end{equation}
where $\mathcal{L}^{std}_{val}(.)$ denotes the validation loss on natural data. Other symbols are the same as Eq. (\ref{eq5}).\par

For a more clear understanding of the proposed two-stage search strategy, an illustration is presented in Figure \ref{two-stage}. The search strategy can be divided into two stages. At the first stage, both the network weights and the architecture parameters are optimized utilizing only the adversarial loss, which we call the normal adversarial training. Once the first-stage optimization is performed a specified number of epochs (e.g., 20 epochs in the figure), the second-stage optimization starts. At the second stage, the network weights are still optimized using the normal adversarial training, while the architecture parameters are optimized using both the adversarial loss and the natural loss, which we call the multi-objective adversarial training. 

It might be questioned that why we do not perform only the second stage optimization to search for accurate and robust neural architectures. Actually, after such an experimental attempt, we find that the validation loss during the search can be optimized smoothly, but the testing performance is not good. This may be the inherent limitation of the differentiable search method, possibly caused by the deep gap \cite{chen2019progressive} and the lack of robustness of the differentiable search algorithm itself \cite{zela2019understanding}. Accordingly, our search strategy may tend to meet this inherent limitation of the differentiable search method. Therefore, the second-stage optimization is only used to fine-tune the architecture. Once the testing performance degrades significantly, we will choose the neural architectures before the performance degradation. We also provide ablation study to demonstrate that the second-stage optimization is beneficial to both the accuracy and the robustness, the details of which are introduced in Subsection \ref{Ablation Study}.
\subsection{Multi-objective Adversarial Training}\label{Multi-objective Adversarial Training}
To address the upper-level multi-objective optimization problem in Eq. (\ref{eq6}), we propose a multi-objective adversarial training method. The pseudo code is presented in Algorithm~\ref{Algorithm_multi}. The method utilizes a gradient-based multi-objective optimization algorithm MGDA \cite{desideri2012multiple,sener2018multi}.\par
\begin{algorithm}
	\SetAlgoLined
	\caption{Multi-Objective Adversarial Training}\label{Algorithm_multi}
	\KwIn{$\omega_{i}$ = network weights after the iteration $i$\\
		\qquad \ \ \ $\alpha_{i-1}$ = architecture parameters after the \\
		\qquad \ \ \ iteration ($i-1$)\\
		\qquad \ \ \ $\mathcal{L}_{val}^{std}(\omega_{i}, \alpha_{i-1})$ = natural loss for optimizing the\\
		\qquad \ \ \ architecture parameters at the iteration $i$\\
		\qquad \ \ \ $\mathcal{L}_{val}^{adv}(\omega_{i}, \alpha_{i-1})$ = adversarial loss for optimizing\\
		\qquad \ \ \ the architecture parameters at the iteration $i$\\
	}
	\KwOut{$\alpha_{i}$ = architecture parameters after the \\
		\qquad \ \ \ iteration $i$}
	Calculate the gradient of the natural loss respect to $\alpha$ by $\theta = \nabla_{\alpha}\mathcal{L}_{val}^{std}(\omega_{i}, \alpha_{i-1})$\\
	Calculate the gradient of the adversarial loss respect to $\alpha$ by $\bar{\theta}=\nabla_{\alpha}\mathcal{L}_{val}^{adv}(\omega_{i}, \alpha_{i-1})$\\
	Determine the weights for the two objectives dynamically by $\gamma^{*}=\underset{0\le\gamma\le1}{argmin}\left\|\gamma \theta+(1-\gamma)\bar{\theta}\right\|^{2}_{2}$\\
	Update $\alpha_{i}$ using $\nabla_{\alpha}( \gamma^{*}\mathcal{L}_{val}^{std}(\omega_{i}, \alpha_{i-1})+(1-\gamma^{*})\nabla_{\alpha}\mathcal{L}_{val}^{adv}(\omega_{i}, \alpha_{i-1})$
\end{algorithm}
MGDA is a gradient-based multi-objective optimization algorithm that either finds a common descent direction for all objectives and performs gradient descent, or does nothing when the current point is Pareto-stationary \cite{desideri2012multiple}. The optimization process of MGDA ensures that any objective will not become worse, so we can use it to construct our multi-objective adversarial training method at the second-stage optimization. The resulting second-stage optimization process can improve the accuracy of the architecture when the robustness is at least better than before conducting the second-stage optimization.\par
Specifically, we first need to determine the weights of all objectives dynamically. The goal is to find a group of weights to transform the multi-objective optimization problem to a single-objective optimization problem, under conditions that optimizing the new single-objective problem is equivalent to optimizing all the objectives simultaneously. Thanks to the two objectives in our method, the process of dynamically determining the weights can be simplified as solving the problem presented in Eq. (\ref{eq7}),
\begin{equation}\label{eq7}
\gamma^{*}=\underset{0\le\gamma\le1}{argmin}\left\|\gamma \theta+(1-\gamma)\bar{\theta}\right\|^{2}_{2}
\end{equation}
where $\theta=\nabla_{\alpha}\mathcal{L}_{val}^{std}(\omega^{*}(\alpha),\alpha)$ and $\bar{\theta}=\nabla_{\alpha}\mathcal{L}_{val}^{adv}(\omega^{*}(\alpha), \alpha)$ denote the gradients of the validation loss on natural data and adversarial examples, respectively, with respect to $\alpha$. Eq.~(\ref{eq7}) has an analytical solution, and can be calculated by Eq.~(\ref{eq8}).
\begin{equation}\label{eq8}
\gamma^{*}=max(min(\frac{(\bar{\theta}-\theta)^{T}\bar{\theta}}{\|\theta-\bar{\theta}\|^{2}_{2}},1),0)
\end{equation}\par
Now the upper-level optimization problem of Eq. (\ref{eq6}) can be transformed to a single-objective optimization problem by calculating the weighted sum of the two objectives using the weights obtained above, which can be formulated as Eq. (\ref{eq9}).
\begin{equation}\label{eq9}
\underset{\alpha}{min}\quad \gamma^{*}\mathcal{L}_{val}^{std}(\omega^{*}(\alpha),\alpha)+(1-\gamma^{*})\mathcal{L}_{val}^{adv}(\omega^{*}(\alpha),\alpha)
\end{equation}\par
Eq. (\ref{eq9}) can then be optimized using gradient decent, which is equivalent to optimizing the original two objectives simultaneously. The optimization process can be formulated as Eq. (\ref{eq10}),
\begin{equation}\label{eq10}
\alpha^{'}=\alpha-\eta_{\alpha}\nabla_{\alpha}( \gamma^{*}\mathcal{L}_{val}^{std}(\omega^{*},\alpha)+(1-\gamma^{*})\mathcal{L}_{val}^{adv}(\omega^{*},\alpha))
\end{equation}
where $\eta_{\alpha}$ is the learning rate for architecture parameters $\alpha$.\par
At the end of search, a discrete neural architecture could be obtained by replacing each mixed operation with the operation that has the largest architecture parameter. The derived neural architecture is intrinsically accurate and robust because we directly optimize the architecture parameters by minimizing the natural loss and the adversarial loss simultaneously.\par
\subsection{Discussion}\label{Method Analysis}
Compared with the existing methods that search for robust neural architectures as mentioned in Section \ref{Introduction}, the proposed method has two significant advantages.\par
Firstly, we successfully apply the adversarial training to the bi-level optimization problem defined by the differentiable search method through alternately updating the network weights and the architecture parameters utilizing the adversarial training. Our method makes full the effectiveness of the adversarial training in minimizing the adversarial loss, so that the searched neural architectures obtain the best robustness among all competitors, which is demonstrated by our experimental results. Meanwhile, owing to the differentiability of the proposed method, we can search for robust neural architectures efficiently. Furthermore, we demonstrate that the adversarial training can be used to train not only the network weights but also the architecture parameters.\par
Secondly, we regard the adversarial training as a multi-objective optimization problem, and explicitly optimize the natural loss and the adversarial loss simultaneously during the search utilizing the proposed multi-objective adversarial training method. The practice makes it possible for the differentiable architecture search method to search for both accurate and robust neural architectures. Although the robustness may be at odds with the accuracy~\cite{tsipras2018robustness,zhang2019theoretically}, we show that we can search out strongly robust neural architectures whose accuracy is also outstanding by making a trade-off between the accuracy and the robustness. Moreover, we notice that the multi-objective adversarial training designed for the architecture parameters is also suitable for the network weights. We provide an idea that the neural networks can be trained to be both accurate and robust through multi-objective optimization, which is worth trying in future works.

\section{Experimental Design}\label{Experimental Design}
In order to evaluate the performance of the proposed DSARA method, we conduct comprehensive experiments to compare it with peer competitors on various benchmark datasets. In this section, we will introduce the experimental settings in detail. To begin with, the peer competitors chosen to compare with the proposed method are introduced in Subsection \ref{Peer Competitors}. Next, the benchmark datasets used by our experiments are shown in Subsection \ref{Benchmark Dataset}. Finally, the parameter settings of the experiments are detailed in Subsection \ref{Parameter Settings}.
\subsection{Peer Competitors}\label{Peer Competitors}
In order to demonstrate the superiority of the proposed DSARA method  accuracy and robustness, the state-of-the-art neural architectures are chosen as the peer competitors from three categories: hand-crafted, standard NAS, and robust NAS.\par
For those neural architectures which are hand-crafted or obtained by standard NAS, they are usually less competitive than the neural architectures obtained by robust NAS when performing the adversarial training, so we simply choose some neural architectures commonly selected by other similar papers. Specifically, the hand-crafted neural architectures chosen includes ResNet-18 \cite{he2016deep} and DenseNet-121 \cite{huang2017densely}, while the neural architectures obtained by standard NAS includes DARTS \cite{liu2018darts} and PDARTS \cite{chen2019progressive}. Besides, the neural architectures obtained by robust NAS are our main competitors, including RobNet \cite{guo2020meets}, AdvRush \cite{mok2021advrush}, RACL~\cite{dong2020adversarially}, \textcolor{COLOR}{DSRNA~\cite{hosseini2021dsrna}, NADAR~\cite{li2021neural},} and ABanditNAS \cite{chen2020anti}. \textcolor{COLOR}{Some of them provide several architectures under different setting. For a fair comparison, we choose the best neural architectures whose number of parameters is similar to most competitors}. Specifically, RobNet searches under different model size and take the cell-free setting as an option, which means all the cells in the neural architecture can optionally have different structures. We choose RobNet-free as our competitor which is searched under the cell-free setting. ABanditNAS architectures are stacked by 6 cells or 10 cells. ABanditNAS-10 refers to the neural architectures searched by stacking 10 cells.
\subsection{Benchmark Dataset}\label{Benchmark Dataset}
The CIFAR-10 \cite{krizhevsky2009learning}, CIFAR-100, SVHN \cite{netzer2011reading} \textcolor{COLOR}{and Tiny-ImageNet-200~\cite{le2015tiny}} benchmark datasets are chosen in the experiments. We choose them for these reasons: 1) they are datasets for the image classification tasks, which are the most critical research field of the adversarial attacks; 2) they are real-world image datasets, which helps to study the adversarial attacks under real conditions; 3) they are the benchmark datasets commonly used by peer competitors.

\textbf{CIFAR:} The CIFAR-10 dataset is a subset of the Tiny Images dataset \cite{torralba200880}, which consists of 60,000 color images, and each is with the dimension of 32 $\times$ 32. 50,000 of the images are divided into the training set, and the remaining 10,000 images are divided into the testing set. The images are categorized into 10 classes, such as frogs and birds. The CIFAR-100 dataset is another subset of the Tiny Images dataset, which has the identical number of images in the training set and the testing set with the CIFAR-10 dataset. The difference is that the images in the CIFAR-100 dataset are divided into 100 categories.\par
\textbf{SVHN:} The SVHN dataset is a real-world image dataset, which comes from house numbers in Google Street View images. The images are of small cropped digits which can be categorized into 0 to 9. The dimension of the images are 32 $\times$ 32. The SVHN dataset consists of more than 600,000 images, including 73,257 digits for training, 26,032 digits for testing, and 531,131 additional to use as extra training data. In our experiments, the additional training data will not be used.

\textcolor{COLOR}{\textbf{Tiny-ImageNet-200:} The Tiny-ImageNet-200 dataset is a subset of the ILSVRC-2012~\cite{russakovsky2015imagenet} classification dataset. It consists of 200 object classes, and for each object class it provides 500 training images, 50 validation images, and 50 test images. All images have been downsampled to 64~$\times$~64~$\times$~3 pixels.}

\subsection{Parameter Settings}\label{Parameter Settings}
\textbf{Search Settings:} Following DARTS, we carry out architecture search on a small network consisting of 8 cells, and hold out half of the CIFAR-10 training data as the validation set. Besides, we set the initial number of channels to be 24, which is larger than conventions, with the purpose of making the model capacity similar to peer competitors for fair comparison, because the model capacity has great impact on robustness \cite{madry2017towards}. Specifically, the numbers of channels divided by the two reduction cells are 24, 48, and 48, respectively. When updating network weights, we always use adversarial examples. When updating architecture parameters, we perform single-objective optimization using only adversarial examples for first 20 epochs, and perform multi-objective optimization using both adversarial examples and natural data for next 10 epochs. To generate adversarial examples for the searching procedure, we use 7-step PGD with the step size of 1/510 and the total perturbation of 2/255. It should be noted that when evaluating the final neural architecture, we perform adversarial training using 7-step PGD with the step size of 2/255 and the total perturbation of 8/255, which is the same as the peer competitors. 

As for other parameters, our settings are the same as DARTS. We use momentum SGD to optimize the network weights $\omega$, with initial learning rate $\eta_{\omega}=0.025$ (annealed down to zero following a cosine schedule), 
momentum 0.9, and weight decay $3\times10^{-4}$. We use Adam as the optimizer for the architecture parameters, with initial learning rate $\eta_{\alpha}=3\times10^{-4}$ for the first-stage single-objective optimization and $\eta_{\alpha}=5\times10^{-5}$ for the second-stage multi-objective optimization, momentum $\beta=(0.5, 0.999)$ and weight decay~$10^{-3}$.\par
\textbf{Evaluation Settings:} To evaluate the searched architecture, we stack 20 cells to form a large network with initial number of channels of 64. Therefore, the number of channels divided by the two reduction cells are 64, 128, and 128, respectively. We perform adversarial training using 7-step PGD with the step size of 0.01 and the total perturbation of 8/255 for 200 epochs. We use SGD to optimize the network, with the momentum of 0.9, and the weight decay of $1\times10^{-4}$. When evaluating on different datasets, the learning rate is a little different, especially the SVHN dataset needs to be trained using a smaller learning rate than CIFAR-10 and CIFAR-100. Specifically, when evaluating on CIFAR-10 and CIFAR-100, the incitial learning rate is set to be 0.1. When evaluating on SVHN, the initial learning rate is set to be 0.01. The learning rate is decayed by the factor of 0.1 at the 100-th and 150-th epoch. The batch size is set to be 32. All the experiments are performed on a single NVIDIA GeForce RTX 2080 Ti GPU card.
\section{Experimental Results and Analysis}\label{Experimental Results and Analysis}
In this section, We first show comprehensive experimental results to analyze the superiority of the proposed method. Specifically, the white-box attack experiments are shown in Subsection \ref{White-box Attacks}, the black-box attack experiments are shown in Subsection \ref{Black-box Attacks}, the transferability experiments are shown in Subsection \ref{Transferability}, and the ablation study for the components of the proposed DSARA method is shown in Subsection \ref{Ablation Study}. Finally, the architectural ingredients of accurate and robust neural networks are analyzed in Subsection~\ref{Architectural Ingredients} to guide both hand-crafting and automatically designing of accurate and robust neural networks.
\begin{figure}[htp]
	\centerline{\includegraphics[width=0.48\textwidth]{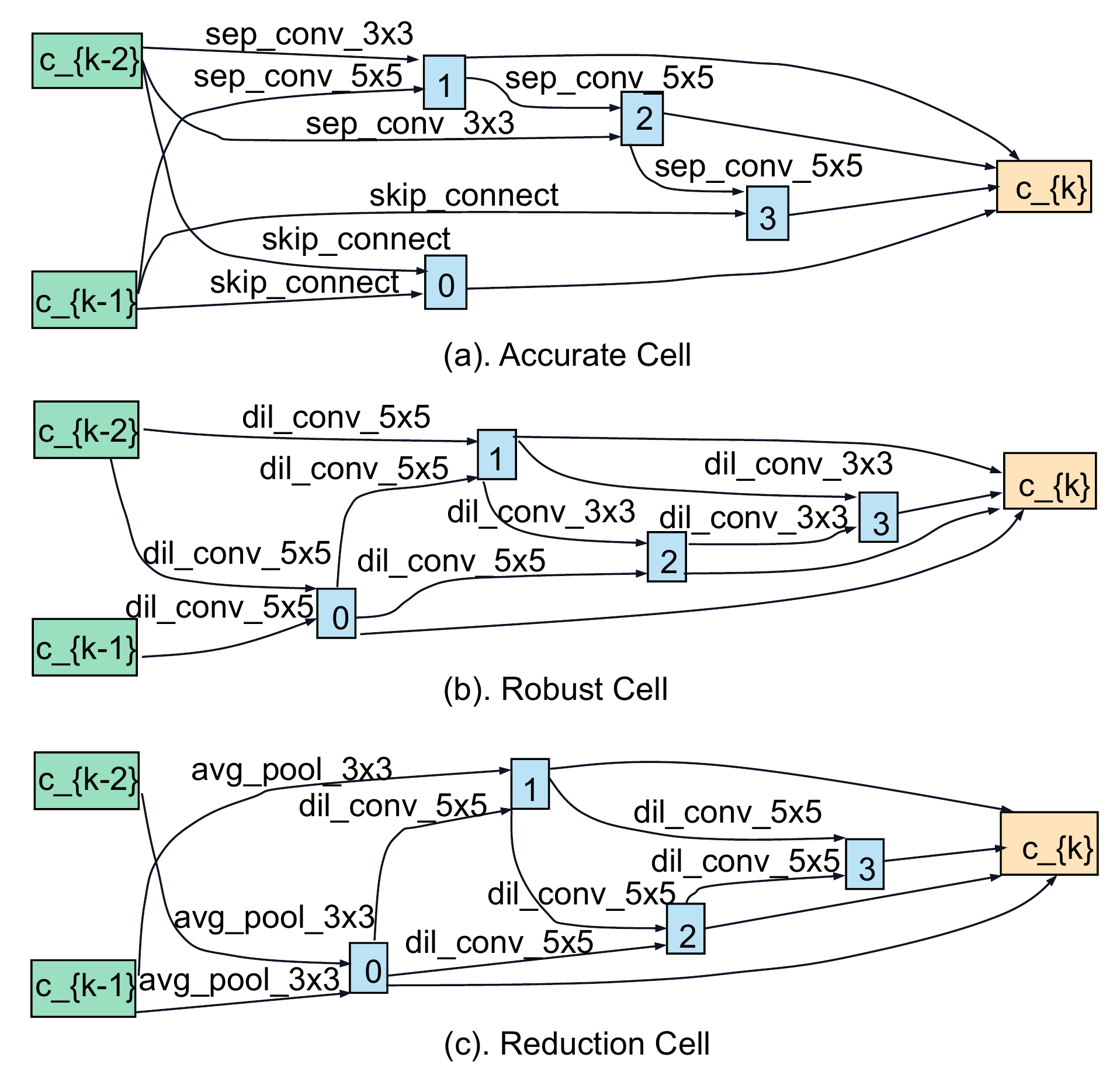}}
	\caption{The visualization of Accurate Cell, Robust Cell and Reduction cell searched by DSARA.}
	\label{cell}
\end{figure}
\begin{table*}
	\caption{Evaluation results \textcolor{COLOR}{\textbf{of adversarially trained models}} on CIFAR-10 under white-box attacks. The best result in each column is in bold, and the second best result is underlined. PGD$^{20}$ and PGD$^{100}$ refer to PGD attack with 20 and 100 iterations, respectively. AA refers to the final evaluation result after completing the standard group of AutoAttack methods. All attacks are $l_{\infty}$-bounded with a total perturbation scale of 8/255.}
	\label{tab1}
	
	\resizebox{\textwidth}{!}{
		\begin{tabular}{c|lcccccccc}
			\toprule
			Category&Model&Params&\textcolor{COLOR}{FLOPs}&Natural Acc.&FGSM&PGD$^{20}$&PGD$^{100}$ & APGD$_{\rm CE}$ & AA\\
			\midrule
			\multirow{3}*{Hand-Crafted}&ResNet-18 & 11.2M &\textcolor{COLOR}{37.67M}& 84.09\% & 54.64\% & 45.86\% & 45.53\% & 44.54\% & 43.22\%\\
			\cmidrule(r){2-10}
			&DenseNet-121 & 7.0M&\textcolor{COLOR}{59.83M}& 85.95\%& 58.46\%& 50.49\%& 49.92\% & 49.11\% & 47.46\%\\
			\midrule
			\multirow{3}*{Standard NAS}&{DARTS} & 3.3M&\textcolor{COLOR}{547.44M}& 85.17\%& 58.74\%& 50.45\%& 49.28\% & 48.32\% & 46.79\%\\
			\cmidrule(r){2-10}
			&PDARTS & 3.4M&\textcolor{COLOR}{550.75M}& 85.37\%& 59.12\%& 51.32\%& 50.91\% & 49.96\% & 48.52\%\\
			\midrule
			\multirow{7}*{Robust NAS}&RobNet-free & 5.6M&\textcolor{COLOR}{800.4M}& 85.00\%& 59.22\%& 52.09\%& 51.14\% & 50.41\% & 48.56\% \\
			\cmidrule(r){2-10}
			&AdvRush & 4.2M&\textcolor{COLOR}{668.53M}& 85.59\%&  59.98\%& 52.76\%&  52.55\% & 51.73\% & 49.28\%\\
			\cmidrule(r){2-10}
			&{RACL} & 3.6M&\textcolor{COLOR}{568.86M}& 83.97\%& 59.29\%& 52.13\%& 51.72\% & 51.24\% & 48.59\% \\
			\cmidrule(r){2-10}
			&{\textcolor{COLOR}{DSRNA}} & \textcolor{COLOR}{2.0M}&\textcolor{COLOR}{336.23M}& \textcolor{COLOR}{80.93\%}& \textcolor{COLOR}{54.49\%}& \textcolor{COLOR}{49.11\%}& \textcolor{COLOR}{48.89\%} & \textcolor{COLOR}{48.54\%} & \textcolor{COLOR}{44.87\%} \\
			\cmidrule(r){2-10}
			&{\textcolor{COLOR}{NADAR}} & \textcolor{COLOR}{4.4M}&\textcolor{COLOR}{700.00M}& \textcolor{COLOR}{\uline{86.23\%}}& \textcolor{COLOR}{60.46\%}& \textcolor{COLOR}{\uline{53.43\%}}& \textcolor{COLOR}{\uline{53.06\%}} & \textcolor{COLOR}{\uline{52.64\%}} & \textcolor{COLOR}{\uline{50.44\%}} \\
			\cmidrule(r){2-10}
			&{ABanditNAS-10} & 5.2M&\textcolor{COLOR}{794.11M}& \textbf{90.64\%}& \textbf{81.31\%} & 50.51\%& 45.73\% & 29.31\% & 16.03\% \\
			\midrule
			Ours& DSARA& 4.5M&\textcolor{COLOR}{1.27G}& 85.92\%& \uline{62.45\%}& \textbf{55.87\%}& \textbf{55.43\%} & \textbf{54.84\%} & \textbf{52.66\%}\\
			\bottomrule
		\end{tabular}
	}
\end{table*}

\subsection{White-box Attacks}\label{White-box Attacks}
\textcolor{COLOR}{The best neural architecture searched by our method is visualized in Figure~\ref{cell}}. We adversarially train the searched architectures and evaluate them under FGSM, PGD$^{20}$, PGD$^{100}$, and the standard group of AutoAttack (APGD$_{\rm CE}$, APGD$^{\rm T}$, FAB$^{\rm T}$, and Square)~\cite{croce2020reliable}. We present the best result obtained by our method, and compare it with some networks which are hand-crafted, derived by standard NAS or derived by robust NAS in Table \ref{tab1}.\par
The results show that the DSARA architecture achieves the highest adversarial accuracy among all competitors under PGD$^{20}$, PGD$^{100}$, and the standard group of AutoAttack, indicating that the DSARA architecture is highly robust. Meanwhile, the natural accuracy of DSARA outperforms all the competitors except DenseNet-121, ABanditNAS-10, and NADAR. Compared with DenseNet-121, the natural accuracy of DSARA is slightly lower (0.03\%) while the accuracy under various attacks is significantly improved (e.g., 5.20\% under AutoAttack). Compared with AbanditNAS-10, DSARA achieves the lower natural accuracy and the lower adversarial accuracy under FGSM attacks, but higher accuracy under PGD attacks and AutoAttack. It is interesting that AbanditNAS-10 only performs well under simple attacks such as FGSM. When the attacks get stronger, AbanditNAS-10 shows lower accuracy than all other competitors. Especially when facing AutoAttack, AbanditNAS-10 can hardly defend the attacks, and the adversarial accuracy of AbanditNAS-10 reduces to only 16.03\%, which indicates that AbanditNAS-10 is not actually trained to be robust. \textcolor{COLOR}{Compared with DSRNA, DSARA achieves better results on both the natural accuracy and the five attacks. Compared with NADAR, DSARA still obtains better results on the five attacks although shows slightly worse performance on the natural accuracy.} Therefore, we come to the conclusion that the DSARA architecture achieves the highest robustness among all competitors while the natural accuracy is also outstanding. In addition, the high accuracy under AutoAttack implies that the DSARA architecture does not unfairly benefit from the obfuscated gradients, because FAB attack in the standard group of AutoAttack is less affected by the obfuscated gradients~\cite{athalye2018obfuscated}.

\begin{table*}
	\captionsetup{width=\textwidth}
	\caption{Evaluation results \textcolor{COLOR}{\textbf{of adversarially trained models}} on CIFAR-10 under transfer-based black-box attacks. Adversarial examples from the source model are generated by PGD$^{20}$ with the total perturbation scale of 8/255.}
	\label{tab2}
	\centering
	\resizebox{0.65\textwidth}{!}{
		\begin{tabular}{l|llll}
			\toprule
			\diagbox[width=80pt]{Source}{Target}&WRN-R&ABanditNAS-10&AdvRush&DSARA\\
			\midrule
			WRN-R&64.76\% & 69.59\% &  68.99\% & 70.03\%\\
			\midrule
			ABanditNAS-10&84.82\% & 50.44\% & 77.58\% & 78.39\%\\
			\midrule
			AdvRush&77.62\% & 68.83\%& 52.76\%& 66.81\%\\
			\midrule
			DSARA&77.09\% & 67.80\%& 64.65\%& 55.87\%\\
			\bottomrule
		\end{tabular}
	}
\end{table*}

\textcolor{COLOR}{In addition, in Table~\ref{tab1}, we have also reported the FLOPs of the models generated by the peer competitors to show their computation cost. As can be seen from Table~\ref{tab1}, the DSARA architecture searched by the proposed method has 1.27G FLOPs, which is significantly more than all other models obtained by the peer competitors, even though their number of parameters are similar to each other. Given that DSARA architecture is more robust, we infer that the number of parameters and the FLOPs are both the factors that affect the robustness of the corresponding architecture. The conclusion also explains why previous studies get totally contradictory conclusions about the effect of model parameters on the robustness, i.e., some studies showed that more parameters can improve adversarial robustness~\cite{xie2019intriguing} while some others showed that more parameters may be harmful to adversarial robustness~\cite{huang2021exploring}. This may be because they ignore the influence of the FLOPs. Moreover, the large FLOPs are essentially caused by the special proportional relationship of the architecture channels designed in the proposed search space, which further demonstrates the effectiveness of the proposed search space. Specifically, the parameter size is proportional to the sum of channel numbers, while the FLOPs are proportional to the product of the channel numbers. The proposed search space keeps the sum of channel numbers similar to conventional search space, but the product of channel numbers is significantly larger, resulting the architecture with similar parameter size but larger FLOPs.}
\subsection{Black-box Attacks}\label{Black-box Attacks}

\begin{table*}
	\captionsetup{width=\textwidth}
	\caption{Evaluation results \textcolor{COLOR}{\textbf{of adversarially trained models}} on CIFAR-100, SVHN, \textcolor{COLOR}{and Tiny-ImageNet} under white-box attacks. PGD$^{20}$ refers to PGD attack with 20 iterations. All attacks are $l_{\infty}$-bounded with a total perturbation scale of 8/255.}
	\label{tab3}
	\centering
	\resizebox{0.7\textwidth}{!}{
		\begin{tabular}{c|l|lll}
			\toprule
			Dataset&Model\qquad\quad\quad&Natural Acc.&FGSM\qquad\qquad&PGD$^{20}$\\
			\midrule
			\multirow{4}*{\quad CIFAR-100\quad}& ResNet-18& 55.57\%& 26.03\%& 21.44\%\\
			\cmidrule(r){2-5}
			&P-DARTS & 58.41\%& 30.35\%& 25.83\%\\
			\cmidrule(r){2-5}
			&DSARA & 58.18\%& 32.60\%& 29.54\%\\
			\midrule
			\multirow{4}*{SVHN}&ResNet-18 & 92.06\%&  88.73\%& 69.51\%\\
			\cmidrule(r){2-5}
			&P-DARTS & 96.02\%& 95.84\%& 87.09\%\\
			\cmidrule(r){2-5}
			&DSARA & 95.84\%& 94.43\%& 92.02\%\\
			\midrule
			\multirow{4}*{\textcolor{COLOR}{Tiny-ImageNet-200}}&\textcolor{COLOR}{WideResNet-34-12} & \textcolor{COLOR}{52.10\%}&  \textcolor{COLOR}{27.82\%}& \textcolor{COLOR}{24.83\%}\\
			\cmidrule(r){2-5}
			&\textcolor{COLOR}{P-DARTS} & \textcolor{COLOR}{42.35\%}& \textcolor{COLOR}{19.74\%}& \textcolor{COLOR}{17.27\%}\\
			\cmidrule(r){2-5}
			&\textcolor{COLOR}{DSARA} & \textcolor{COLOR}{40.62\%}& \textcolor{COLOR}{18.99\%}& \textcolor{COLOR}{17.61\%}\\
			\bottomrule
		\end{tabular}
	}
\end{table*}
Due to the competitive performance of AdvRush and ABanditNAS-10, we further conduct black-box attacks among their models and ours. We conduct transfer-based black-box attacks, attacking the target model using adversarial examples generated by the source model. Besides, we perform extended experiments on WRN-34-R, which is the most robust variant of WideResNet found by [43]. WRN-34-R is trained using additional 500k data, so it shows the highest accuracy on both natural images and adversarial examples. We are interested in how DSARA behaves when facing such a highly robust model. Black-box evaluation results are presented in Table \ref{tab2}.\par
The results show that DSARA is more resilient against transfer-based black-box attacks than AdvRush and ABanditNAS. For example, when considering the model pair AdvRush $\leftrightarrow$ DSARA, AdvRush $\rightarrow$ DSARA achieves the attack success rate (i.e., 100\% - adversarial accuracy) of 33.19\%, while DSARA $\rightarrow$ AdvRush achieves the attack success rate of 35.35\%, where there is a gap of 2.16\%. When considering the model pair ABanditNAS-10 $\leftrightarrow$ DSARA, ABanditNAS-10 $\rightarrow$ DSARA achieves the attack success rate of 21.61\%, while DSARA $\rightarrow$ ABanditNAS-10 achieves the attack success rate of 32.20\%, where there is a gap of 10.59\%. When facing the highly robust model WRN-R, DSARA also shows higher attack success rate than AdvRush and AbanditNAS-10 (22.91\% compared with 22.38\% and 15.18\%) and higher prediction accuracy (70.03\% compared with 68.99\% and 69.59\%). In conclusion, the black-box evaluation results further demonstrate the highest robustness of the DSARA architecture among competitors. Meanwhile, it proves again that the DSARA architecture does not unfairly benefit from the obfuscated gradients because the transfer-based black-box attacks do not use the gradients of the target model.
\subsection{Transferability to Other Datasets}\label{Transferability}
To demonstrate the transferability of the DSARA architecture, we transfer the architecture searched on CIFAR-10 to CIFAR-100, SVHN, \textcolor{COLOR}{and Tiny-ImageNet-200.} The evaluation results are presented in Table \ref{tab3}.

\textcolor{COLOR}{Firstly, the analysis on the small-scale datasets (i.e., CIFAR-100 and SVHN in the table) is carried out.} Compared with the hand-crafted baseline model ResNet-18, DSARA performs better on both natural images and adversarial examples no matter which dataset they are transferred to. Compared with the standard NAS baseline model P-DARTS, DSARA achieves significantly higher accuracy under PGD$^{20}$ (3.71\% higher on CIFAR-100, and 4.93\% higher on SVHN), while the natural accuracy reduction can be ignored (0.23\% lower on CIFAR-100, and 0.18\% lower on SVHN). The results show that the DSARA architecture is still highly robust after being transferred to other \textcolor{COLOR}{small-scale} datasets while its accuracy is also outstanding.

\textcolor{COLOR}{Secondly, the analysis on the large-scale dataset (i.e., Tiny-ImageNet-200 in the table) is made. In the case of the Tiny-ImageNet-200 dataset, the architectures searched by NAS algorithms are compared with WideResNet-34-12~\cite{zagoruyko2016wide} (which is a powerful hand-crafted architecture on the ImageNet dataset) instead of ResNet-18. As can be seen in the table, DSARA performs similarly to P-DARTS, with slightly lower natural accuracy (1.73\% lower) and FGSM accuracy (0.75\% lower) but higher PGD$^{20}$ accuracy (0.34\% higher). However, DSARA and P-DARTS are both significantly worse than WideResNet-34-12, no matter compared through natural accuracy (more than 9\% lower) or adversarial accuracy (more than 8\% lower under FGSM and more than 7\% lower under PGD$^{20}$). The results confirm that NAS-based architectures searched on the CIFAR-10 dataset cannot transfer well to the Tiny-ImageNet-200 dataset in the case of adversarial training.}

\textcolor{COLOR}{Existing works showed that architectures searched by NAS algorithms are more robust for small-scale datasets and simple tasks than hand-crafted architectures, but the opposite is true when the dataset size or the task complexity increases~\cite{huang2021exploring,devaguptapu2021adversarial}. The above experiments are just consistent with the observations.}

\subsection{Ablation Study}\label{Ablation Study}
\textbf{Search Space Settings.} We try all possible placement of the Accurate Cells and the Robust Cells while the placement of the Reduction Cells is the same as the NASNet search space. Because the Accurate Cell and the Robust Cell are the same except the placement which results in different cell structures, the case that employs two Robust and one Accurate Cell is equivalent to the case that employs one Robust Cell and two Accurate Cells, and we just need to experiment on one of the two cases. When the placement is determined, we further study on the different settings of the number of filters. We only make a preliminary exploration when changing the proportional relationship of the number of filters after the second Reduction Cell. \par
\begin{figure*}
	\includegraphics[width=\textwidth]{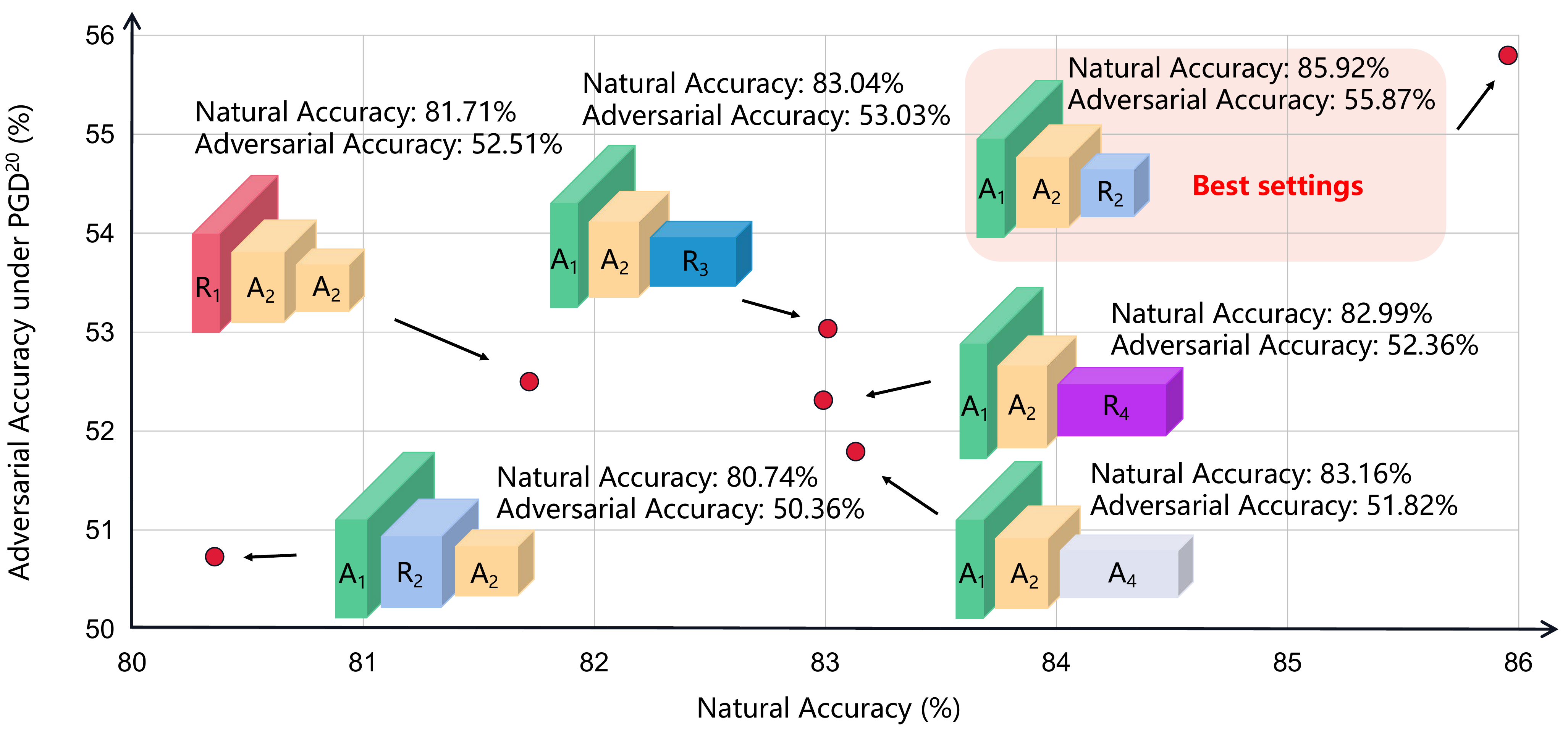}
	\caption{Visualization analysis of neural architectures shown in Table \ref{tab4} under different placement settings and filter settings. In this figure, character R refers to the Robust Cell, character A refers to the Accurate Cell, and the subscript of A and R refers to the filter settings. For example, R$_{4}$ refers to the Robust Cell with the number of filters four times the initial.}\label{search_space_compare}
\end{figure*}
\begin{table}
	\captionsetup{width=0.45\textwidth}
	\caption{Ablation study on CIFAR-10 for search space settings.}
	\label{tab4}
	\centering
	\resizebox{0.45\textwidth}{!}{
		\begin{tabular}{ccc|cr}
			\toprule
			\textcolor{COLOR}{Row Number}&Placement&Filter Setting&Natural Acc.&PGD$^{20}$\\
			\midrule
			\textcolor{COLOR}{1}&A-A-R & 1-2-2 & 85.92\% & 55.87\%  \\
			\midrule
			\textcolor{COLOR}{2}&R-A-A & 1-2-2 & 81.71\% & 52.51\% \\
			\midrule
			\textcolor{COLOR}{3}&A-R-A & 1-2-2 & 80.36\%& 50.74\%\\
			\midrule
			\textcolor{COLOR}{4}&A-A-R & 1-2-3& 83.04\%& 53.03\% \\
			\midrule
			\textcolor{COLOR}{5}&A-A-R & 1-2-4& 82.99\%& 52.36\%\\
			\midrule
			\textcolor{COLOR}{6}&A-A-A & 1-2-4& 83.16\%& 51.82\%\\
			\bottomrule
		\end{tabular}
	}
\end{table}

We search under different settings and retrain the searched neural architectures. The results on CIFAR-10 are shown in Table \ref{tab4}. To represent the placement of the cells, we use A for Accurate Cell and R for Robust Cell. For example, A-A-R means we place the Accurate Cells before the second Reduction Cell and the Robust Cells after the second Reduction Cell, which is the same as the DSARA search space described in Subsection \ref{Search Space}. For filter settings, we use N$_{1}$-N$_{2}$-N$_{3}$ to represent the proportional relationship of the number of filters. For example, 1-2-4 means the number of filters between the first Reduction Cell and the second Reduction Cell is two times the initial number of filters, and the number of filters after the second Reduction Cell is four times the initial number of filters, which is adopted by the NASNet search space.\par
When the filter setting is fixed to be 1-2-2 \textcolor{COLOR}{(experiments 1, 2 and 3)}, the best result is achieved when we set the placement to be F-F-R \textcolor{COLOR}{(experiment 1)}. So we fix the placement to be F-F-R and conduct further experiments. When the placement is fixed to be F-F-R \textcolor{COLOR}{(experiments 1, 4 and 5)}, the best result is achieved exactly when we set the number of filters to be 1-2-2 \textcolor{COLOR}{(experiment 1)}, which is adopted in the previous experiments. Therefore, we construct the final DSARA search space with the placement of F-F-R and the filter setting of 1-2-2. Compared with conventional search space (placement of F-F-F and filter setting of 1-2-4), the neural architecture searched from our search space achieves 2.23\% higher natural accuracy and 2.78\% higher PGD$^{20}$ accuracy. \par
We also provide a visualization analysis in Figure \ref{search_space_compare} for the above experimental results. As shown in the figure, an inappropriate setting may lead to the low accuracy and the low robustness of the searched neural architectures (bottom left corner of the figure), while a well-designed setting chosen by our method can largely improve the accuracy and the robustness upper bound (top right corner of the figure). \textcolor{COLOR}{Please note that the results shown in Figure~\ref{search_space_compare} and Table~\ref{tab4} denote the solution searched from different search space, which lies in different Pareto Fronts, i.e., they do not imply that accuracy and robustness are not two competing objectives.}\par

\begin{table*}
	\caption{Ablation study on CIFAR-10 for critical components of DSARA.}
	\label{tab5}
	\centering
	\resizebox{1.0\textwidth}{!}{
		\begin{tabular}{c|l|cccc|cc}
			\toprule
			\textcolor{COLOR}{Row Number}&Method &Adv. Train &DSARA Search Space&Two-Stage Search Strategy&\textcolor{COLOR}{The Proposed Multi-Objective} &Natural Acc.& PGD$^{20}$\\	
			\midrule
			\textcolor{COLOR}{1}&RACL & \XSolidBrush & \XSolidBrush & \XSolidBrush & \textcolor{COLOR}{-} &83.89\% & 49.34\%  \\
			\midrule
			\textcolor{COLOR}{2}&AdvRush & \XSolidBrush & \XSolidBrush & \XSolidBrush & \textcolor{COLOR}{-}&85.59\% & 52.76\%\\
			\midrule
			\textcolor{COLOR}{3}&\multirow{8}*{DSARA}&\CheckmarkBold & \XSolidBrush & \XSolidBrush&\textcolor{COLOR}{-}& 83.51\% & 53.98\% \\
			\cmidrule(r){3-8}
			\textcolor{COLOR}{4}&& \CheckmarkBold& \CheckmarkBold& \XSolidBrush & \textcolor{COLOR}{-}&85.23\% & 53.99\% \\
			\cmidrule(r){3-8}
			\textcolor{COLOR}{5}&& \CheckmarkBold& \XSolidBrush& \CheckmarkBold & \textcolor{COLOR}{\CheckmarkBold}&83.16\% & 51.82\% \\
			\cmidrule(r){3-8}
			\textcolor{COLOR}{6}&& \textcolor{COLOR}{\CheckmarkBold}& \textcolor{COLOR}{\CheckmarkBold}& \textcolor{COLOR}{\CheckmarkBold} & \textcolor{COLOR}{\XSolidBrush}&\textcolor{COLOR}{85.04\%} & \textcolor{COLOR}{53.72\%} \\
			\cmidrule(r){3-8}
			\textcolor{COLOR}{7}&& \CheckmarkBold& \CheckmarkBold& \CheckmarkBold & \textcolor{COLOR}{\CheckmarkBold}&85.92\% & 55.87\% \\
			\bottomrule
		\end{tabular}
	}
\end{table*}
\textbf{Contributions of Different Components of the DSARA Method.} We evaluate the contributions made by three important components of the proposed DSARA method, including adversarial training during the search, the DSARA search space, and the two-stage search strategy. The results are summarized in Table \ref{tab5}. In the table, the option `Adv. Train' refers to whether the overall architecture is optimized utilizing the adversarial training during the search or not. The proposed DSARA method must use the adversarial training during the search while other competitors such as RACL and AdvRush utilizes some indirect metrics as mentioned before. The option `DSARA Search Space' refers to whether the search is conducted on the proposed DSARA search space or the NASNet search space. The option `Two-Stage Search Strategy' refers to whether the search is performed by the proposed two-stage search strategy or only the first-stage single-objective optimization. \textcolor{COLOR}{The option `The Proposed Multi-Objective' refers to whether the proposed multi-objective adversarial training method is used or it is replaced with the objective function constructed by simply adding the natural loss and the adversarial loss. When the two-stage search strategy is not applied, the option `The Proposed Multi-Objective' is not considered.}

As can be seen in Table~\ref{tab5}, when only performing adversarial training to search for robust neural architectures \textcolor{COLOR}{(the third row, which is equal to applying adversarial training to the searching process of DARTS)}, the architecture found by DSARA shows higher robustness than RACL \textcolor{COLOR}{(the first row)} and AdvRush \textcolor{COLOR}{(the second row)}, which demonstrates the adversarial training is not only a theoretically feasible method, but also a practically feasible method that can be applied to the differentiable search method to search for the most robust architectures. The method is simple but effective, which lacks practice and comparison in previous works. When the adversarial training is performed during the search, we further study on the remaining two components in the proposed DSARA method. Compared with only performing the adversarial training during the search \textcolor{COLOR}{(the third row)}, if we employ the DSARA search space while not employing the proposed two-stage search strategy \textcolor{COLOR}{(the fourth row)}, both the natural accuracy and accuracy increases, indicating that architectures from the DSARA search space improves both the accuracy and the robustness upper bound of the searched neural architectures. If we further employ the two-stage optimization method \textcolor{COLOR}{(the seventh row)}, both the natural accuracy and the PGD$^{20}$ accuracy reach the highest, which demonstrates that the two-stage search strategy helps to find both accurate and robust architectures. We also notice that when only performing the two-stage optimization without using the DSARA search space \textcolor{COLOR}{(the fifth row)}, the result is worse than the architecture searched by performing single-objective optimization \textcolor{COLOR}{(the third row)}, which may indicate that the two-stage optimization must be combined with the DSARA search space. \textcolor{COLOR}{To figure out the effect of the proposed multi-objective adversarial training method, we further perform the corresponding ablation study by replacing the multi-objective adversarial training method with another multi-objective method (i.e., the one summing the natural loss and the adversarial loss as one objective), and the results are shown in the sixth row of Table~\ref{tab5}. As can be seen in Table~\ref{tab5}, when using the sum of the natural loss and the adversarial loss (the sixth row), both the natural accuracy and the adversarial accuracy decrease compared with the use of the proposed multi-objective adversarial training method (the seventh row). This proves the superiority of the proposed multi-objective adversarial training method in terms of both accuracy and robustness.}

\subsection{\textcolor{COLOR}{Analysis on Architectural Ingredients of Accurate and Robust Neural Networks}}\label{Architectural Ingredients}

\textcolor{COLOR}{To get more insights of the robustness, we further analyze the experimental results.}

\section{Conclusion and Future Work}\label{Conclusion}
In this work, we propose the DSARA method to search for accurate and robust neural architectures after adversarial training automatically and efficiently. Specifically, we first design the DSARA search space specially for adversarial training through experiments, which empirically improve the accuracy and the robustness of the searched neural architectures. Then we design a gradient-based two-stage search strategy, searching for accurate and robust neural architectures efficiently. In particular, at the first stage, we update the architecture to minimize the adversarial loss, which makes full use of the effectiveness of the adversarial training in enhancing the robustness and leads to the stronger robustness of the searched neural architectures than previous works. At the second stage, the multi-objective adversarial training is proposed to optimize the architecture to be both accurate and robust by minimizing the natural loss and the adversarial loss simultaneously. In spite of the conflict between the accuracy and the robustness, we demonstrate that we can make a trade-off between the accuracy and the robustness to find both accurate and robust neural architectures. We evaluate the searched architecture under various adversarial attacks on various benchmark datasets, which demonstrates the effectiveness of our method to search for both accurate and robust architectures.

Besides, by analyzing the search results of our method, we reach an important conclusion that the cell structures in different positions of the neural architectures play different roles for the accuracy and the robustness, and the neural architectures can obtain the higher accuracy and the higher robustness simultaneously by deploying very different cell structures in different positions. We notice that although it is widely used to construct neural architectures by repeated stacking, few people realize the importance of stacking different structures in different positions. We believe our conclusion will have great guiding significance on both hand-crafting and automatically designing of accurate and robust neural architectures. In the future, we plan to research on why the structures in different positions of the neural architectures have different impacts on the accuracy and the robustness to further guide the design of neural architectures theoretically.

\normalem
\bibliographystyle{plain}
\bibliography{ref.bib}

\end{document}